\documentclass[sigconf]{acmart}
\usepackage[ruled,vlined]{algorithm2e}
\usepackage{titlesec}
\usepackage{amsmath}
\usepackage{mathtools}
\usepackage{graphicx}
\usepackage{adjustbox}

\DeclareMathOperator*{\argmax}{arg\,max}
\DeclarePairedDelimiter\ceil{\lceil}{\rceil}
\DeclareMathOperator{\Tr}{Tr}

\usepackage{xcolor}
\definecolor{purple}{HTML}{9F4C7C}

\titleformat{\subsubsection}
  {\normalfont\normalsize\itshape}{\thesubsubsection}{1em}{}
\titlespacing*{\subsubsection}{0pt}{3.25ex plus 1ex minus .2ex}{0ex plus .2ex}

\AtBeginDocument{%
  \providecommand\BibTeX{{%
    \normalfont B\kern-0.5em{\scshape i\kern-0.25em b}\kern-0.8em\TeX}}}

\setcopyright{acmcopyright}
\copyrightyear{2020}
\acmYear{2020}
\acmDOI{}

\acmConference[Paper under review]{}{}{}
\acmBooktitle{}
\acmPrice{}
\acmISBN{}

\begin{document}

\title{GRAFFL: Gradient-free Federated Learning of a Bayesian Generative Model}

\author{Seok-Ju Hahn, Junghye Lee}
\email{sjhahn11512@unist.ac.kr, junghyelee@unist.ac.kr}
\orcid{}
\affiliation{%
  \institution{Management Engineering, Ulsan National Institutue of Science and Technology (UNIST)}
  \streetaddress{}
  \city{}
  \state{}
  \postcode{}
}

\renewcommand{\shortauthors}{}


\begin{abstract}

Federated learning platforms are gaining popularity. One of the major benefits is to mitigate the privacy risks as the learning of algorithms can be achieved without collecting or sharing data. While federated learning (i.e., many based on stochastic gradient algorithms) has shown great promise, there are still many challenging problems in protecting privacy, especially during the process of gradients update and exchange. This paper presents the first gradient-free federated learning framework called GRAFFL for learning a Bayesian generative model based on approximate Bayesian computation. Unlike conventional federated learning algorithms based on gradients, our framework does not require to disassemble a model (i.e., to linear components) or to perturb data (or encryption of data for aggregation) to preserve privacy. Instead, this framework uses implicit information derived from each participating institution to learn posterior distributions of parameters. The implicit information is summary statistics derived from SuffiAE that is a neural network developed in this study to create compressed and linearly separable representations thereby protecting sensitive information from leakage. As a sufficient dimensionality reduction technique, this is proved to provide sufficient summary statistics. We propose the GRAFFL-based Bayesian Gaussian mixture model to serve as a proof-of-concept of the framework. Using several datasets, we demonstrated the feasibility and usefulness of our model in terms of privacy protection and prediction performance (i.e., close to an ideal setting). The trained model as a quasi-global model can generate informative samples involving information from other institutions and enhances data analysis of each institution.

\end{abstract}



\maketitle

\section{INTRODUCTION}
As privacy and security have become important issues, federated learning (FL) is being noticeable as a remarkable solution of machine learning (ML) in recent years. FL is an emerging configuration of ML techniques, where collaborative learning of a global model between individuals or institutions is possible with no migration of data under the administration of a central server (CS). Since it helps to protect sensitive information while increasing the utility of data analysis, FL has been positioning as a legitimate reliever for ML in a situation that sensitive information is involved. Mainly two situations --- horizontally or vertically distributed data --- are considered in FL: the former is the case where the distributed samples share equivalent features while the latter is one of overlapped samples having distinct features. There has been a wealth of research, but many challenges still remain open in how to design the FL setting to enhance the potential of ML methods towards meaningful discoveries, the solution of which requires novel techniques in many fields.

Existing FL settings are usually placed on sharing gradient or weight information during a model training phase \cite{S+15, M+16}; utilization of the derivative information is usually powerful since it guarantees to find an optimal search direction of steepest ascent. However, these methods should be judicious since methods and analyses on gradient-driven privacy leakage has been actively investigated \cite{Sh+17, Hi+17, Me+19, Zh+19, Pa+19, Na+19}. For example, a model can unintentionally memorize training input for generalization \cite{Zh+16, Ca+19} with derivative information produced during a model training phase.



A dominant class of methods for secure and privacy-preserving FL is to exchange noise added/encrypted gradients/weight updates of a local model with the CS by using novel techniques such as differential privacy (DP) \cite{Ge+17, Ab+16, Ch+09}, homomorphic encryption (HE) \cite{Yu+14, Gi+18, Ha+11}, and multi-party computation (MPC) \cite{Yu+14, Gi+18, Ha+11}.
Such studies \cite{Kim+18, K+19} have been successfully conducted, but some limitations have been reported including computational complexity and accuracy loss of the model. DP, which ensures a strong privacy, only applies to the output of the algorithm, which might introduce too many noises in cases where the transfer or exchange of intermediate statistics (including the output), such as distributed learning, is required. HE and MPC, on the other hand, are computationally intensive and often require an approximation even for simple computations \cite{Ao+16, Ki+18, Ya+19}.





While gradient information is the key to learn a model in existing FL algorithms, it is sometimes showing limited capabilities in real practices since the target function to be optimized is sometimes non-convex, or analytically intractable (e.g. due to latent variables, or complicated integration in a likelihood term) \cite{Ru+13, Bu+07, TV12}. When if the objective function has multiple optima, the gradient-based methods can miss the global one \cite{Ru16}. Above all, it can also be an incipience or a target of attacks \cite{Zh+19, S+15}. 

It is, therefore, conceivable to consider of gradient-free methods in FL frameworks. Formerly, there has been an attempt to introduce a gradient-free distributed optimization algorithm \cite{Wa+19}, but it is not in accord with FL scheme as they did not consider preserving privacy, and it is not completely gradient-free as they partially applied Kiefer-Wolfowitz algorithm for approximating gradients. In a similar manner, \cite{Ch+19} focused on a parallelization of model training with no concern on preserving privacy, and utilized synthetic gradients. Heretofore, no gradient-free FL in a true sense exists to the best of our knowledge. 


In this research, we at first introduce a novel FL scheme to rule out the source of gradients: a likelihood. Likelihood-free inference methods can be an useful alternative in FL as a gradient-free method as they make use of function values instead of derivative information for finding optimal set of parameters. 
More specifically, Approximate Bayesian computation (ABC) \cite{RD+84} will be explored in this study: a Bayesian parameter inference method that only requires a parameterized generative model and prior distributions for simulating data and parameters respectively \cite{TV+12}; data generated from the model of which parameters given by prior distributions are compared with the observed data in the form of summary statistics. Although the use of summary statistics is encouraged in ABC to avoid the curse of dimensionality that occurs in calculating distances, the challenge lies in how to construct "sufficient" summary statistics beyond informative summary statistics \cite{FP12,B+13}. The main contributions of this work are as follows:


\vspace{-4mm}
\begin{itemize}
\item \textbf{The first gradient-free FL framework:} 
As a new alternative to gradient-based FL algorithms, we develop the \pmb{gra}dient-\pmb{f}ree \pmb{f}ederated \pmb{l}earning framework (GRAFFL), aiming to train a federated generative model in a horizontally distributed setting. Instead of locally training models having the pre-agreed structure and then sharing updated gradients to CS, our method aims to train a global model at the CS in likelihood-free manner with no exposing of model structure and no exchange of gradients or parameters with local parties. Thereby, each local site has flexibility in a construction of own predictive models after finishing FL training phase, which encourages fine-tuned analyses suitable for situations of each site.


\item \textbf{New sufficient dimensionality reduction for sufficient summary statistics:}
We propose a new technique called SuffiAE based on learning of the modified Auto-Encoder (AE) to construct sufficient summary statistics and theoretically prove that this yields sufficient summary statistics via sufficient dimensionality reduction (SDR).

\item \textbf{FL of a Bayesian generative model:} To serve the proof of the concept of GRAFFL, we use the Gaussian mixture model (GMM) as the parameterized generative model, which is recognized as the most representative and successful technique. The GMM trained by our framework can have parameters estimated by leveraging prior information provided by domain experts and statistical evidence from data across multiple institutions in an efficient way. We demonstrate its feasibility through a few experiments including a comparison of classification performance in different scenarios after simulating a situation where imbalanced and insufficient samples derived from different modes of multi-modal population distribution are distributed to each site.


\end{itemize}

\section{PRELIMINARIES}
\subsection{Approximate Bayesian Computation}
ABC explores a parameter space by sampling from prior distribution, and then collect reasonable parameters by comparing summary statistics of observed data and simulated data from parameters using the discrepancy metric with the acceptance threshold value. This allows us to estimate the posterior distributions of the parameters in place of directly optimizing the likelihood term. 


\subsubsection{Summary statistics}
Utilizing raw data itself may cause computational inefficiency, the curse of dimensionality, and the threat of information leakage. Therefore, it is important to construct proper summary statistics such that (1) sufficient and (2) reducing a dimensionality of data \cite{AE+12, GP13}.
Instead of using simple sufficient statistics such as a mean or a variance, utilizing well-established dimensionality reduction techniques such as a hash function, AE, and principal component analysis enables ABC parameter estimation to be robust even in high-dimensional data \cite{IZ+19, Pr15, Su+13}.

\begin{definition}
\textbf{Bayes sufficiency} \cite{KO42} \\
For given data $\pmb{\mathrm{x}} \in \mathbb{R}^{D}$, summary statistics $\mathrm{S}(\mathbf{\pmb{\mathrm{x}}})$ derived from a transformation $\mathrm{S(\cdot)}: \mathbb{R}^{D} \rightarrow \mathbb{R}^{d}$ is sufficient if it satisfies
\begin{equation}\notag
p(\pmb{\uptheta} | \mathbf{\pmb{\mathrm{x}}})=p(\pmb{\uptheta} | \mathrm{S}(\mathbf{\pmb{\mathrm{x}}}))
\end{equation}
for all prior $p(\pmb{\uptheta})$.
\end{definition}

\begin{definition}
\textbf{Sufficient dimensionality reduction} adapted from \cite{WC07} \\ 
In a typical regression setting, where a univariate response variable $Y$ is estimated on $\mathbf{X}$ with a function $f(\cdot)$, i.e. $\textnormal{E}[Y|\mathbf{X}]=f(\mathbf{X})$, dimensionality reduction mapping $\mathrm{S(\cdot)}: \mathbb{R}^{D} \rightarrow \mathbb{R}^{d}, d \leq D$ is \textit{sufficient} if the following condition is met:
\begin{equation}\notag
Y \perp \mathbf{X} | S(\mathbf{X}).    
\end{equation}
It implies that $\textnormal{E}[Y|\mathbf{X}]=\textnormal{E}[Y|S(\mathbf{X})]$ if $\mathrm{S(\cdot)}$ is a transformation satisfying SDR.
\end{definition}

In summary, when we use a sufficient summary statistics, it is guaranteed that the approximated posterior derived through ABC is converged to the true posterior estimated from the original data $\pmb{\mathrm{x}}$. In addition, this trait may also help to accomplish data anonymization in our framework as the reduced data is usually indecipherable.
\subsubsection{Acceptance threshold}
Selection of an appropriate threshold value $\epsilon$ is essential to make ABC rejection sampling approximately converge to the true posterior.
In fact, ABC rejection sampling presented in Algorithm~\ref{alg:ABCRS} is one of the simplest ABC parameter estimation methods \cite{Ta+97, Pr+99}.
There is a trade-off between computational cost and approximation accuracy in estimating parameters \cite{B+02}.
With an appropriate discrepancy metric $\operatorname{\Delta}\left(\cdot, \cdot\right)$, sufficient summary statistics $\mathrm{S(\cdot)}$, and adequately small $\epsilon$, the joint posterior is derived as:
\begin{equation} \notag
p_{\epsilon}(\pmb{\uptheta}, \pmb{\mathrm{x}}_{\text{gen}} | S(\pmb{\mathrm{x}}_{\text{obs}}))
=\frac{p(\pmb{\uptheta}) p(\pmb{\mathrm{x}}_{\text{gen}} | \pmb{\uptheta}) \mathbb{I}_{\epsilon}(\pmb{\mathrm{x}}_{\text{gen}})}{\int_{\Theta}\int_{\chi} p(\pmb{\uptheta}) p(\pmb{\mathrm{x}}_{\text{gen}} | \pmb{\uptheta}) \mathbb{I}_{\epsilon}(\pmb{\mathrm{x}}_{\text{gen}}) \mathrm{d} \pmb{\mathrm{x}}_{\text{gen}} \mathrm{d} \pmb{\uptheta}}
\end{equation}
for the sample space $\chi$ and the parameter space $\Theta$, where $\mathbb{I}_{\epsilon}(\pmb{\mathrm{x}}_{\text{gen}})=[\operatorname{\Delta}\left(S(\pmb{\mathrm{x}}_{\text{gen}}), S(\pmb{\mathrm{x}}_{\text{obs}})\right)\leq \epsilon]$. 
Assuming sufficient summary statistics and a small $\epsilon$, then the true posterior is approximated by:
\begin{equation} \notag
p_{\epsilon}(\pmb{\uptheta} | S(\pmb{\mathrm{x}}_{\text{obs}}))
= \int_{\chi} p_{\epsilon}(\pmb{\uptheta}, \pmb{\mathrm{x}}_{\text{gen}} | S(\pmb{\mathrm{x}}_{\text{obs}})) \mathrm{d} \pmb{\mathrm{x}}_{\text{gen}}.
\end{equation}
The accuracy of an approximation is highly contingent upon the choice of an acceptance value. Its effect on the approximation of the true posterior is stated as:
\begin{equation}\notag
\lim_{\epsilon \rightarrow 0} p_{\epsilon}(\pmb{\uptheta} | S(\pmb{\mathrm{x}}_{\text{obs}})) \cong p(\pmb{\uptheta} | \mathbf{\pmb{\mathrm{x}}}), 
\lim_{\epsilon \rightarrow \infty} p_{\epsilon}(\pmb{\uptheta} | {S}(\pmb{\mathrm{x}}_{\text{obs}}))\cong p(\pmb{\uptheta}). 
\end{equation}
In other words, a large threshold value learns nothing for the posterior, which eventually converges to the prior while a small threshold value approximately approaches the posterior.
A discrepancy metric $\operatorname{\Delta}\left(\cdot, \cdot\right)$ (or a similarity measure) is required for comparing the distance between generated samples $\pmb{\mathrm{x}}_{\text{gen}}$ and observed data $\pmb{\mathrm{x}}_{\text{obs}}$.

Meanwhile, another condition is required for samples to be generated from the approximately true posterior. When we accept samples satisfying $[\pmb{\mathrm{x}}_{\text{gen}}=\pmb{\mathrm{x}}_{\text{obs}}]=1$, where $[\cdot, \cdot]$ is the Iverson bracket notation, it requires a long time to accept samples, even in a discrete sample space \cite{We16}. 

Therefore, the acceptance of sample proposals is determined with a proper discrepancy metric and a small enough acceptance threshold value.
Conventionally, the distance metric defined in the sample space $\chi$ (e.g. Euclidean distance) is used as the discrepancy metric \cite{TV+12}.


\begin{algorithm}
	\caption{ABC Rejection Sampling}
	\label{alg:ABCRS}
    \underline{\text{Input}} \newline
    $L: \text{maximum number of accepted parameters}$  \newline
    $N: \text{total number of samples and parameter proposals}$ \newline
    $\pmb{\mathrm{x}}^{\textit{obs}}: \text{observed data} \in \mathbb{R}^{D}$ \newline
    ${S}(\pmb{\mathrm{x}}): \text{sufficient summary statistics of } \pmb{\mathrm{x}} \in \mathbb{R}^{d} (d \leq D)$ \newline
    $p(\pmb{\uptheta}): \text{prior of parameter(s) } {\theta}_{i} \in \mathbb{R}^{p}, (i=1,...,N)$ \newline
    $\operatorname{\Delta}\left(\cdot, \cdot\right): \text{discrepancy metric } \mathbb{R}^{d} \rightarrow \mathbb{R}$ \newline
    $\epsilon: \text{acceptance threshold}$
    
    \underline{\text{Output}} \newline
    $p_{\epsilon}(\pmb{\uptheta} | {S}(\mathbf{\pmb{\mathrm{x}}}^{\textit{obs}})): \text{a posterior distribution}$
    
    \underline{\text{Pseudocode}} \newline
        Set $L$ and $N$ $(L \leq N)$ \newline
	    Set $\#params = 0$ \newline
	    Compute ${S}(\pmb{\mathrm{x}}^{\textit{obs}}):=\pmb{\mathrm{x}}^{\textit{enc}}$ \newline
		\While{$\#params < L$}{
        Simulate $N$ parameters from prior: ${\theta}^{*}_{i} \sim p(\pmb{\uptheta})$.
        
        Simulate $\pmb{\mathrm{x}}^{\textit{gen}}_{i} \in \mathbb{R}^{d}$ samples from the simulation model: $\pmb{\mathrm{x}}^{\textit{gen}}_{i} \sim p(\cdot|{\theta}_{i}^{*})$

          \If{$\operatorname{\Delta}(\pmb{\mathrm{x}}^{\textit{enc}}, \pmb{\mathrm{x}}^{\text{gen}}_{i}) < \epsilon$}{
           Store ${\theta}^{*}_{i}$
           
           $\#params \gets \#params + 1$
          }
         }
\end{algorithm}

\subsection{Bayesian Gaussian Mixture Model}
\label{sec:BGMM}
A GMM is a parameterized probabilistic function for estimating an unknown arbitrary probability density, assuming that all data points are from the finite mixture of Gaussian distributions. In other words, the unknown density function can be represented as the weighted sum of Gaussian components.

Suppose we have an input matrix $\pmb{\mathbf{X}}_{\textit{obs}}\in\mathbb{R}^{M \times D}$, of which row corresponds to a sample vector $\pmb{\mathbf{x}}_{\textit{obs}, i} \in \mathbb{R}^{D} (i=1, \ldots, M)$, and the input data are from the mixture of $K$ Gaussian distributions. 
GMM typically consists of three parameters, $\pmb{\uptheta}=\left\{\pi_{k}, \pmb{\mu}_{k}, \pmb{\mathbf{\Sigma}}_{k}\right\}_{k=1}^{K}$; $\pi_{k}$ is a responsibility vector indicating the probability that the input data belong to the $k$-th latent cluster. 
The latent cluster can be modeled by a latent random variable $\pmb{\mathbf{z}}=\left\{z_{k} | z_{k} \in {\{0,1\}}\right\}_{k=1}^{K}$, of which distribution is parameterized by $\pmb{\uppi}=\left\{\pi_{k} | 0 \leq \pi_{k} \leq 1\right\}_{k=1}^{K}$, where $p\left(z_{k}=1\right) =\pi_{k}$.
The input data assigned to any latent cluster $k$ follow normal distribution with mean $\pmb{\mu}_{k}$ and covariance $\pmb{\mathbf{\Sigma}}_{k}$. Then, the density of input data is estimated by GMM as
$ p(\pmb{\mathbf{x}})=\sum_{\pmb{\mathbf{z}}} p(\pmb{\mathbf{z}}) p(\pmb{\mathbf{x}} | \pmb{\mathbf{z}})=\sum_{k=1}^{K} \pi_{k} \mathcal{N}\left(\pmb{\mathbf{x}} | \pmb{\mu}_{k}, \pmb{\mathbf{\Sigma}}_{k}\right) $, 
where
$p(\pmb{\mathbf{z}})=\prod_{k=1}^{K} \pi_{k}^{z_{k}}, p(\pmb{\mathbf{x}} | \pmb{\mathbf{z}}) =\prod_{k=1}^{K} \mathcal{N}\left(\pmb{\mathbf{x}} | \pmb{\mu}_{k}, \pmb{\mathbf{\Sigma}}_{k}\right)^{z_{k}}$, 
and $\sum_{k} z_{k}=1, \sum_{k} \pi_{k}=1$. 
\begin{figure*}[t]
  \centering
  \includegraphics[width=0.90\textwidth]{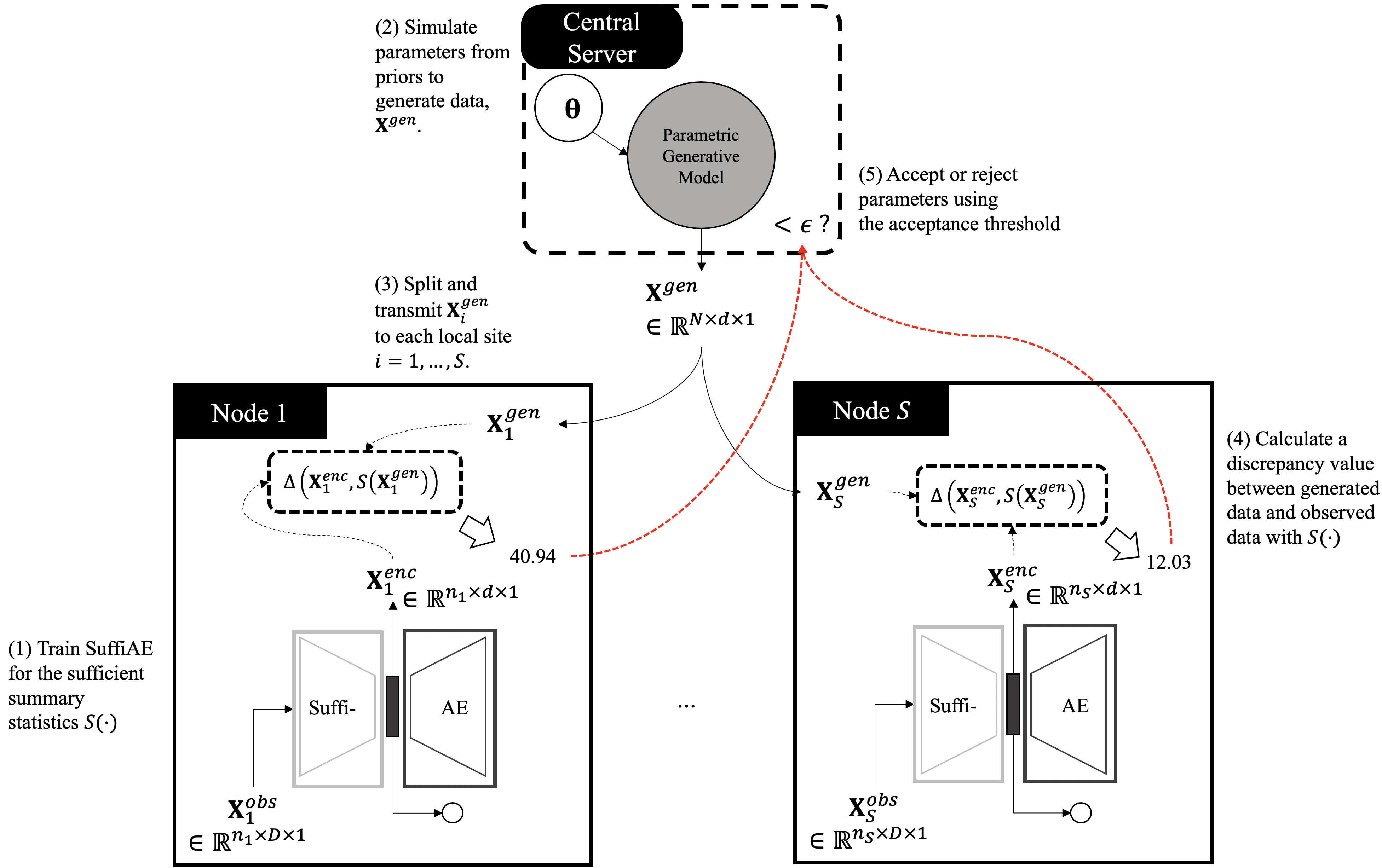}
  \caption{GRAFFL Framework}
  \label{fig:GRAFFL}
\end{figure*}
For the Bayesian estimation, we need to introduce a suitable prior distribution of each parameter \cite{Bi06}. 
For $\pmb{\uppi}$, a proper prior distribution is known as a Dirichlet distribution: $\pmb{\uppi} \sim \operatorname{Dir}(\pmb{\uppi} | \pmb{\upalpha})$, where $\pmb{\upalpha}=\left(\alpha_{1}, \alpha_{2}, \ldots, \alpha_{K}\right)$. 
When $K$ is determined, the prior distribution of $\pmb{\uppi}$ regards $K$ as the number of categories.
Next, for $\pmb{\mathbf{\upmu}}=\left\{\pmb{\mu}_{k} | \pmb{\mu}_{k} \in \mathbb{R}^{D}\right\}_{k=1}^{K}$ and $\pmb{\mathbf{\Sigma}}=\left\{\pmb{\mathbf{\Sigma}}_{k} | \pmb{\mathbf{\Sigma}}_{k} \in \mathbb{R}^{D \times D}\right\}_{k=1}^{K}$, their prior distributions are intertwined; so-called Normal-Inverse-Wishart distribution.
At first, a covariance matrix $\pmb{\mathbf{\Sigma}}_{k}$ is sampled from Inverse-Wishart prior distribution: $\pmb{\mathbf{\Sigma}}_{k}\sim\mathcal{W}^{-1}\left(\pmb{\mathbf{\Sigma}}_{k} | \nu, \pmb{\mathbf{\upPsi}}\right)$, which is parameterized by a positive-definite scale matrix $\pmb{\mathbf{\upPsi}} \in$ $\mathbb{R}^{D \times D}$, and a degree of freedom, $\nu > D-1$.
Then, using the sampled covariance matrix $\pmb{\mathbf{\Sigma}}_{k}$, the prior distribution of $\pmb{\mu}_{k}$, a multivariate-normal distribution becomes available: $\pmb{\mu}_{k} \sim \mathcal{N}\left(\pmb{\mu}_{k} | \pmb{m}, \frac{1}{\kappa} \pmb{\mathbf{\Sigma}}_{k}\right)$. A location vector $\pmb{m} \in \mathbb{R}^{D}$ and a positive real number $\kappa$ are required.


\section{GRAFFL: Gradient-free federated learning framework}
GRAFFL aims to work in a situation where each participating site has a biased distribution of data distinct from others by learning a generative model that can implicitly reflect sample diversities in all participating sites without derivative information.
The proposed framework is illustrated in Figure~\ref{fig:GRAFFL}.
The main assumptions of our framework are: (1) all local parties have horizontally distributed data with different densities, and they want to train own model in a supervised manner. 
(2) Each participant is reluctant to share or open their local data even if the data is transformed or anonymized.
(3) The CS is always semi-honest and curious and there always is a possibility of data leakage by the CS.
These are usually knotty problems in gradient-based ordinary FL, but GRAFFL can easily handle those with the ABC scheme.

Before describing the general process of GRAFFL, we define a few notations.
The difference between generated samples to be transferred to site $j$ ($\pmb{\mathbf{X}}^{\textit{gen}}_{j}=\{\pmb{\mathbf{x}}^{\textit{gen}}_{j,k} \in \mathbb{R}^{d}\}_{k=1}^{n_j} \in \mathbb{R}^{n_j \times d}$) and local samples at site $j$ transformed into sufficient summary statistics ($S\big(\pmb{\mathbf{X}}^{\textit{obs}}_{j}\big):=\pmb{\mathbf{X}}^{\textit{enc}}_{j}=\{\pmb{\mathbf{x}}^{\textit{enc}}_{j,k} \in \mathbb{R}^{d}\}_{k=1}^{n_j} \in \mathbb{R}^{n_j \times d}$) is measured with the predefined discrepancy metric, and dimensions of generated data should be equal to dimensions of anonymized data in each site (i.e., $d$). 
In this study, the Euclidean distance is used as a discrepancy metric $\operatorname{\Delta}\left(\cdot, \cdot\right)$, and the discrepancy is calculated in a batch $\operatorname{\pmb{\Delta}}_{j}$:
\begin{equation}\notag
\operatorname{\pmb{\Delta}}_{j}:=\Big\{\operatorname{{\Delta}}\Big(\pmb{\mathbf{X}}^{\textit{enc}}_{j},\pmb{\mathbf{X}}^{\textit{gen}}_{j}\Big)=\left\|\pmb{\mathbf{x}}^{\textit{enc}}_{j,k}
-\pmb{\mathbf{x}}^{\textit{gen}}_{j,k}\right\|_{2}^{2}\Big|{k=1,...,n_j}\Big\}
\end{equation}
where $j$ is an index indicating each local site ($j=1,...,S$), $n_{j}$ is the number of samples of each site.

The process of estimating the posterior of a generative model's parameters in the CS is as follows.
\begin{itemize}
\item Each local party trains its own SuffiAE (presented in section~\ref{sec:Suffi}) for generating summary statistics of which dimensionality is $d$. At the same time, each local party should agree on which discrepancy metric they will use. 
\item The CS determines the total number of parameter proposals, $N$, the maximum number of accepted parameters, $L$, and proper prior distributions for each parameter, $p(\pmb{\uptheta})$.
\item The CS simulates $N$ parameters from priors and generate samples $\pmb{\mathbf{X}}^{\textit{gen}}_{j}$ from the proposed parameters. Then, samples are split by $n_j$, and are sent to each corresponding local site $j$. i.e. total number of samples transmitted to local sites is $\sum_{j}{n_j}$. Therefore, total iterations are $\ceil*{\frac{N}{\sum_{j}{n_j}}}$.
\item At each local site, delivered samples $\pmb{\mathbf{X}}^{\textit{gen}}_{j}$ are compared with summary statistics of each local samples $\pmb{\mathbf{X}}^{\textit{enc}}_{j}$ using predefined discrepancy metric $\operatorname{\Delta}\left(\cdot, \cdot\right)$. Then, set of calculated discrepancy  $\operatorname{\pmb{\Delta}}_{j}$ is sent back to the CS.
\item The CS receives and collects the aggregated discrepancy and sort them in an ascending order. With iterations are proceeded, the order should also be updated.
\item After all iterations, $L$ parameters generating $L$ smallest $\pmb{\mathbf{X}}^{\textit{gen}}_{j}$ is accepted. The largest discrepancy among $L$ smallest generated samples is to be the threshold $\epsilon$.
\end{itemize} 

The overall algorithm with the GMM specifically chosen as a parametric generative model is presented in \textbf{Algorithm 2}. Note that this is a proof of the concept version of GRAFFL.
The posteriors of parameters of GMM are to be estimated at the CS by exchanging minimal information with local sites. 
The information is aggregated discrepancy values between generated samples from the model in the CS and modified samples in each site. In other words, the only information transmitted out of local sites to the CS is the discrepancy value calculated in each local site.
Once the inference of posterior distributions is completed, plausible samples with diminished dimension $d$ can be generated at the CS, and then are sent to each site for supplementing skewed data distribution of each local site.
An AE \cite{Ru+86} is one type of neural networks learned in a self-supervised manner by minimizing the reconstruction error between original input and reconstructed input from encoded data.
A basic AE is composed of two sub-networks of a symmetric structure: an encoder network and a decoder network.
The former is to shrink the input vector into a lower dimensional vector, and the latter is to reconstruct the original input from the compressed vector.
The AE is mainly used to derive latent representations by reducing the dimensions of the input feature space while removing noise inherent in the data. 
The objective function of AE is usually provided as minimizing $\left\|\pmb{\mathbf{x}}-\pmb{\mathbf{x}}^{\prime}\right\|_{2}^{2}$
, where the reconstructed input is $\pmb{\mathbf{x}}^{\prime}=h_{\text {dec}}\left(\pmb{\mathbf{W}}_{\text {dec}}\left(h_{\text{enc}}\left(\pmb{\mathbf{W}}_{\text {enc}} \pmb{\mathbf{x}}+{b}_{\text{enc}}\right)\right)+{b}_{\text{dec}}\right)$
and $h(\cdot),\pmb{\mathbf{W}},b$ denote an element-wise activation function, a weight matrix, and a bias term of encoder and decoder network, respectively.

In this research, we modify the structure of AE to develop SuffiAE for accomplishing purposes of (1) generating sufficient summary statistics for ABC, which compensates weakness of ABC on high-dimensional data, and (2) preserving privacy in data by transforming through serial non-linear mappings in the encoder network.

\subsection{SuffiAE: Auto-Encoder for sufficient dimensionality reduction and data anonymization}\label{sec:Suffi}
\begin{figure*}
  \centering
  \includegraphics[width=0.8\linewidth]{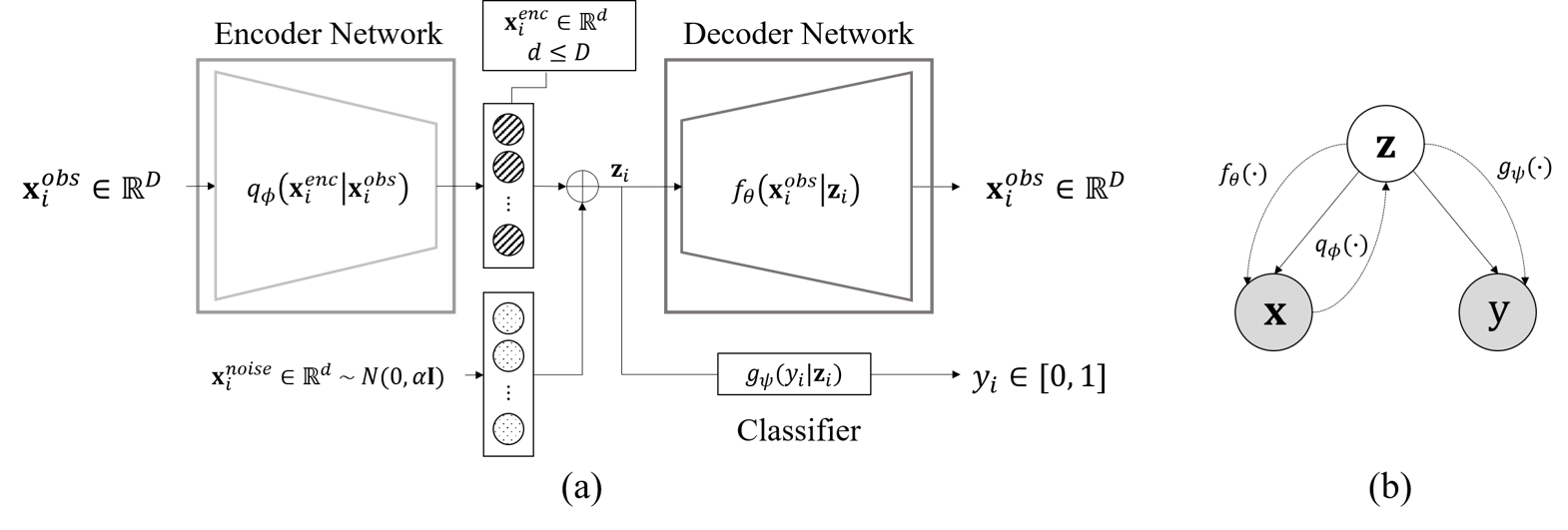}
  \caption{(a) Configuration and (b) Graphical Representation of SuffiAE}
  \label{fig:Suffi}
\end{figure*}

For (1), we adopted the structure of \cite{Ba+18}, and a proof of sufficiency will be given in section 3.2 and Appendix A.
For (2), it makes sense to use diminished data in lieu of original ones for the protection of sensitive information. A class of methods for protecting data securely is to transform data space as a means of data anonymization by utilizing irreversible operations (i.e., a way of circumventing privacy protection) such as one-way function \cite{Yu+06, Le+18, Di+18, Bo+11}, nonlinear transformation \cite{Bh+10, Ba+04, Li+06, Ol+04}; SuffiAE belongs to both. 

SuffiAE is trained independently in each site: no need to agree on the structure of SuffiAE between local parties, no need to share the structure or trained parameters during the training phase. 
The pact reached by sites before their local learning of SuffiAE includes the dimensionality of the modified output and the composition and the ordering of input features. To achieve the latter, private set intersection \cite{Ch+17} or secure alignment of feature modes proposed by \cite{Kim+17} should be applied beforehand at each site.
Since a deeper neural network architecture is advantageous \cite{Go16}, more than a single layer can be considered for configurations of encoder and decoder networks.
For satisfying sufficiency of encoded output $\pmb{\mathrm{z}}_{i}$, additional noise term $\pmb{\mathrm{x}}^{\textit{noise}}_{i}$ sampled from $\mathcal{N}\left(\pmb{0}, \upalpha \pmb{\mathbf{I}} \right)$ with an arbitrary small constant ${\upalpha}$ is added to the encoded vector $\pmb{\mathrm{x}}^{\textit{enc}}_{i}$, i.e. $\pmb{\mathrm{z}}_{i}=\pmb{\mathrm{x}}^{\textit{noise}}_{i}+\pmb{\mathrm{x}}^{\textit{enc}}_{i}$ for (1) injecting stochastic property and (2) calibrating errors induced by the acceptance threshold $\epsilon$ (refer to noisy ABC \cite{FP12}).

While achieving its original purpose, the AE at the same time optimizes compressed representations to be linearly separable by inserting a bypath to encoded output sites and adding a new objective function. 
Thereby, SuffiAE is able to (1) generate a low dimensional vector that can be well-classified by a linear model (see Figure~\ref{fig:Suffi}-(a)), and (2) satisfy characteristics of sufficient dimensionality reduction defined in $\textit{Definition 2.2}$. 

The objective function of SuffiAE at each site $j, (j=1, ..., S)$ is defined as:
\begin{multline}\notag
\text{maximize } \sum_{i=1}^{n_{j}}\Bigg[-\frac{1}{2}\|\pmb{\mathbf{x}}_{i}-f_{\theta}(\pmb{\mathbf{z}}_{i})\|_{2}^{2}+\Big\{y_{i} \log g_{\psi}(\pmb{\mathbf{z}}_{i})  \\
+(1-y_{i}) \log (1-g_{\psi}(\pmb{\mathbf{z}}_{i}))\Big\} -\frac{1}{2}\Big\{\|q_{\phi}(\pmb{\mathbf{x}}_{i})\|_{2}^{2}+d(\alpha-1-\log \alpha)\Big\}\Bigg]
\end{multline}
, where $\pmb{\mathbf{x}}_{i} \text{ and } \pmb{\mathbf{z}}_{i}$ represent the $i$-th original vector, a latent vector respectively, and the corresponding $y_{i} \in\{0,1\}$ indicates one of true labels in a local training set.
$f_{\theta}(\cdot)$, $g_{\psi}(\cdot)$, and $q_{\phi}(\cdot)$ are the mapping functions learned through the decoder network, the encoder network, and the classifier, respectively.

\subsection{Analyses on GRAFFL}
\textbf{i) Sufficiency of summary statistics driven by SuffiAE}
According to the definition of sufficient dimensionality reduction (SDR) defined in $\textit{Definition 2.2}$, we need to find a mapping $S(\cdot)$ that can retain all critical information about $\pmb{\mathrm{x}}$. 

It is equivalent to find a latent variable $\pmb{\mathrm{z}}$ represented in figure~\ref{fig:Suffi}-(b) \cite{Ba+18}.
In other words, a deterministic and differentiable mapping $q_{\phi}(\cdot)$ is used as a sufficient dimensionality reduction function.
However, since it is difficult to directly maximize a likelihood $p(\pmb{\mathrm{x}}, \mathrm{y})$ because we used a neural network structure, which has many parameters with non-linear mappings, we instead aim to find a variational lower bound and maximize this bound.

Our objective is to find a set of parameters $(\theta^{*}, \psi^{*}, \phi^{*})$.
\begin{equation*}
(\theta^{*}, \psi^{*}, \phi^{*})=\argmax_{\theta, \psi, \phi}{\mathcal{L}(\pmb{\mathrm{x}}, \mathrm{y})}
\end{equation*}
where
\begin{multline*}
{\mathcal{L}(\pmb{\mathrm{x}}, \mathrm{y})}=\log{p(\pmb{\mathrm{x}}, \mathrm{y})} \geq \\
E_{\pmb{\mathrm{z}} \sim q_{\phi}(\pmb{\mathrm{z}}|\pmb{\mathrm{x}})}[\log{p(\pmb{\mathrm{x}}|f_{\theta}(\pmb{\mathrm{z}}))}]+
E_{\pmb{\mathrm{z}} \sim q_{\phi}(\pmb{\mathrm{z}}|\pmb{\mathrm{x}})}[\log{p(\mathrm{y}|g_{\psi}(\pmb{\mathrm{z}}))}]\\ -\operatorname{KL}[q_{\phi}(\pmb{\mathrm{z}}|\pmb{\mathrm{x}}) || p(\pmb{\mathrm{z}})].
\end{multline*}

The prior distribution $p(\pmb{\mathrm{z}})$ is assumed to follow Gaussian distribution with zero mean and identity variance as proposed in \cite{KW14}.
This is stated as Eq (1), being adopted in our setting. See Appendix A for a detailed proof.

\textbf{ii) Choice of an acceptance threshold}
Adjusting proper threshold is difficult, but research on robust threshold decision is still at its infancy; just determined by a domain expert.
Some theoretical analyses on the relationship between an acceptance threshold value and an asymptotic convergence property of ABC exist \cite{Bi+15, Fa+13, Ba+15};, it is limited to be applied in a practical setting due to many assumptions that are hardly satisfied in reality.

Though we adopted a perspective proposed in \cite{Bi+15}, which treats the term which compares a discrepancy with a threshold, \\ i.e. $\operatorname{\Delta}(\pmb{\mathrm{x}}^{\textit{enc}}, \pmb{\mathrm{x}}^{\text{gen}}) < \epsilon$, as finding a k-nearest $L$ samples in the ball of radius $\epsilon$, of which center is the observed sample, $\pmb{\mathrm{x}}^{\textit{obs}}$. \\ i.e. $\mathcal{B}_{d}\left(\pmb{\mathrm{x}}^{\textit{obs}}, \epsilon\right)=\left\{\pmb{\mathrm{x}} \in \mathbb{R}^{d}:\left\|\pmb{\mathrm{x}}^{\textit{gen}}-\pmb{\mathrm{x}}^{\textit{obs}}\right\| \leq \epsilon\right\}$.

Instead of defining a threshold value $\epsilon$ above all, we instead accumulate discrepancy value calculated at each local site, and then order corresponding samples into an increasing order.
\begin{equation*}\notag
\pmb{\mathrm{X}}^{\textit{gen}}=\pmb{\mathrm{X}}^{\textit{gen}}_{\text{accepted}} \cup \pmb{\mathrm{X}}^{\textit{gen}}_{\text{rejected}}= \\ 
\Big\{\pmb{\mathrm{x}}^{\textit{gen}}_{(1)}, ..., \pmb{\mathrm{x}}^{\textit{gen}}_{(L)}\Big\} \cup \Big\{\pmb{\mathrm{x}}^{\textit{gen}}_{(L+1)}, ..., \pmb{\mathrm{x}}^{\textit{gen}}_{(N)}\Big\}
\end{equation*}
Then, the acceptance threshold value can be decided as:
\begin{equation}\notag
\epsilon = \inf_{\pmb{\mathrm{x}} \in \pmb{\mathrm{X}}^{\textit{gen}}_{\text{rejected}}}{\operatorname{\Delta}(\pmb{\mathrm{x}}^{\textit{enc}}, \pmb{\mathrm{x}})}
\end{equation}

This approach is also useful for the attack scenario (will be discussed in 3.4.ii) where some local sites intentionally transmits inflated discrepancies to the CS: it is automatically filtered out since the discrepancy value is ordered at the CS. i.e. an abnormally large discrepancy value is to be assigned to $\pmb{\mathrm{X}}^{\textit{gen}}_{\text{rejected}}$ with high probability.

\textbf{iii) Equivalence of ABC in a distributed and centralized setting}
The main difference between \textbf{Algorithm 1} (centralized ABC) and \textbf{Algorithm 2} (quasi-distributed ABC) is the part of processing discrepancy values.
GRAFFL does not transmit the global model to be learned to each site, and no necessity for doing so, because learning separated data without aggregation is possible by modifying small part of the original algorithm. It is the reason why GRAFFL is considered as \textit{quasi-distributed} version of ABC.

For centralized ABC, calculating discrepancy values between generated samples and observed samples can be done at once. For GRAFFL, difference values are evaluated at each site independently with mutually exclusive samples and then aggregated.
Without loss of generality, the two process can be said to be equivalent when if the order of observed samples is not shuffled in the centralized case.

\begin{algorithm}
	\caption{GRAFFL for Bayesian GMM}
    \underline{\text{Input}} \newline
    $n_{j}: \text{number of training data in site } j \ (j=1,...,S)$\newline
    $\pmb{\mathbf{X}}^{\textit{obs}}_{j} \in \mathbb{R}^{n_{j} \times D}: \text{data at local party }j$ \newline
    ${\mathbf{y}}^{\textit{obs}}_{j} \in \mathbb{R}^{n_{j}}: \text{corresponding label at local party }j$ \newline
    $N: \text{Total number of sample and parameter proposals}$ \newline
    $L: \text{maximum number of accepted parameters}$ \newline
    $S: \text{number of local parties}$ \newline
    $d: \text{diminished dimension}$ \newline
    $\text{SuffiAE}_{j}: \text{SuffiAE at each local party }j$ $\text{(returns encoder network } Enc_{j}(\cdot),\newline\text{for generating latent representations }\pmb{\mathbf{X}}^{\textit{enc}}_{j} \in \mathbb{R}^{n_{j} \times d})$ \newline
    $\#epochs: \text{number of epochs to train the SuffiAE}$ \newline
    $\operatorname{\Delta}(\cdot, \cdot): \text{discrepancy metric defined in 3.1}$ \newline
    $\epsilon: \text{an acceptance threshold value}$ \newline
    $\operatorname{Dir}(\pmb{\uppi}|\pmb{\upalpha}): \text{prior of }\pmb{\uppi}$ \newline
    $\operatorname{NIW}(\pmb{\upmu}, \pmb{\Sigma}|\nu, \pmb{\upPsi}, \pmb{m}, \kappa): \text{prior of }\pmb{\pmb{\upmu}, \pmb{\Sigma}}$ \newline
    \underline{\text{Output}} \newline
    $p_{\epsilon}\Big(\pmb{\uptheta} | \sum_{j=1}^{S}\pmb{\mathbf{X}}^{\textit{enc}}_{j}\Big): \text{posterior distributions of the set of parameters }\newline\pmb{\uptheta}=\{\pmb{\uppi}, \pmb{\upmu}, \pmb{\Sigma}\}$ \newline
    \underline{\text{Pseudocode}} \newline
    \centerline{\text{/* At each local party */}} \newline
    Filter common features and aligning orders using PSI \newline
    Consent on diminished dimension $d$ and transmit this information to the CS \newline
    \centerline{\text{/* At CS */}}
    Set a maximum number of accepted parameters $L$ \newline
    Set an appropriate acceptance threshold value $\epsilon$ \newline
    Construct a discrepancy metric $\operatorname{\Delta}(\cdot, \cdot)$ and send it to each local party \newline
	\centerline{\text{/* At each local party */}} \newline
	    \For{$j \text{ in } S$}{
	        \For{$\_ \text{ in } \#epochs$}{
	            Train $\text{SuffiAE}_{j}$
	            
	            Generate latent representations $\pmb{\mathbf{X}}^{\textit{enc}}_{j}=Enc_{j}\Big(\pmb{\mathbf{X}}^{\textit{obs}}_{j}\Big) \in \mathbb{R}^{n_{j} \times d}$
	        }
        }
		$\#iteration = 0$
		
		\While{$\#iteration < \ceil*{\frac{N}{\sum_{j}{n_j}}}$}{
		\centerline{\text{/* At CS */}}
        Simulate set of $N$ parameters $\pmb{\uptheta}^{*}=\{\pmb{\uppi}^{*}, \pmb{\upmu}^{*}, \pmb{\Sigma}^{*}\}$ from each prior distribution: $\pmb{\uppi}^{*} \sim \operatorname{Dir}(\pmb{\uppi}|\pmb{\upalpha})$, $(\pmb{\upmu}^{*}, \pmb{\Sigma}^{*} ) \sim \operatorname{NIW}(\pmb{\upmu},\pmb{\Sigma}|\nu, \pmb{\upPsi}, \pmb{m}, \kappa)$
        
        Generate data $\pmb{\mathrm{X}}^{\textit{gen}} \sim p(\cdot|\pmb{\uptheta}^{*})$
        
        Split $\pmb{\mathrm{X}}^{\textit{gen}}$ according to $n_{j}$; 
        $\pmb{\mathrm{X}}^{\textit{gen}}_{j} \in \mathbb{R}^{n_{j} \times d}$
        
        Send the copy of $\pmb{\mathrm{X}}^{\textit{gen}}_{j}$ to each site $j$
        \centerline{\text{/* At each local party */}}
        
        Calculate $\operatorname{\pmb{\Delta}}_{j}:=\operatorname{\Delta}\Big(\pmb{\mathrm{X}}^{\textit{enc}}_{j}, \pmb{\mathrm{X}}^{\textit{gen}}_{j}\Big)$
        
        Send the value $\operatorname{\pmb{\Delta}}_{j}$ to the CS
        
        \centerline{\text{/* At CS */}} 
        Receive and collect $\operatorname{\pmb{\Delta}}_{j}$
        
        Order $\pmb{\mathrm{X}}^{\textit{gen}}_{j}$ in an ascending order corresponding to $\operatorname{\pmb{\Delta}}_{j}$ and store them in $\pmb{\mathrm{X}}^{\textit{gen}}$
        
        $\#iteration \gets \#iteration + 1$
    }
    \centerline{\text{/* At CS */}}
    Accept first $L$ parameters in $\pmb{\mathrm{X}}^{\textit{gen}}$
\end{algorithm}

\subsection{Privacy analysis}
\textbf{i) Semi-honest (or fully-dishonest) and curious CS}
If the CS intends to retrieve one of local parties data, the CS should know the dimension and ordering of original features, some amount of original data, and optionally weights and structure of decoder network in the target local party. However, except the situation when the target participant is willing to give the original data, this cannot be accomplished. 
It should be noted that the latent representations generated from the SuffiAE cannot be recovered to their original input if each local party does not disclose their trained decoder networks weights and structure to the public, thereby data can be preserved.

\textbf{ii) Collusion between CS and one or more local parties against a specific party}
It is still impossible to make a leakage in a specific local participant even in the situation that one or more participants cooperate with the CS.
It attributed to auto-associative property of Auto Encoder \cite{Al+12}. Even the CS or other parties succeed in skimming other site's data, it is still a compressed representation of the original data; this cannot be fully recovered using decoder network of other party. In other words, recovering decreased data using other participant's decoder network yields samples with low-confidence.

\textbf{iii) Collusion between one or more local parties against a specific party}
It is more difficult to collude among some local parties except the CS, because the only information that a local party and the CS exchange is a discrepancy value, but it is not shared among parties.
Therefore, without the CS, it is very tough to snatch information of target party. Even though the confederates manage to intercept the information of target party (similarity values), it is almost impossible to fully retrieve the original data due to the same reason of the second scenario. 

\textbf{iv) Advantages of using SuffiAE}
Since the trained weights and the structure of SuffiAE at each site are never opened to outside, and original samples are never used after finishing a training of SuffiAE at each site, a robust protection of privacy can be achieved.
Even though summarized samples are exposed to out of a local site, it is almost not available to retrieve the original date. As it is a typical ill-posed problem to find a mapping from data in $\mathbb{R}^{d}$ to $\mathbb{R}^{D}$, where $d \leq D$ \cite{Ka08}, it is complicated for attackers to retrieve the original data from transformed ones unless the attacker has information on the weight and structure of decoder network and some portion of original data.
Moreover, what reinforces secureness is the fact that each local party has a flexibility in organizing a structure of SuffiAE; even if a structure of SuffiAE in one party is revealed by mistake, further accidents can be prevented.

\section{EXPERIMENTS}
\subsection{Overview}
We performed a few experiments to confirm that (1) the proposed gradient-free method can learn distributed information successfully, and thus (2) it can boost further analyses in each local site (by alleviating data imbalance issue and data scarceness issue).

\subsection{Capability of learning distributed information}
\textbf{i) Experimental set-up} We configured a simple set-up for verifying that GRAFFL framework can train a model in a horizontally distributed manner. A two-dimensional tri-modal distribution of which component is a mixture of three Gaussian distributions is considered as a population distribution. Each single Gaussian distribution is: $\mathcal{N}\left(\pmb{\mathrm{\mu}}_{k}, \pmb{\mathbf{I}}\right)$, where $\pmb{\mathrm{\mu}}_{1} = [-9, 3]^\intercal, \pmb{\mathrm{\mu}}_{2} = [0, -9]^\intercal, \pmb{\mathrm{\mu}}_{3} = [8, 3]^\intercal$, and its responsibility parameter $\pmb{\mathrm{\pi}}=[\frac{1}{3},\frac{1}{3},\frac{1}{3}]^\intercal$. 
The choice of hyperparameters of prior distributions, i.e. $\pmb{\upalpha}, \nu, \pmb{\upPsi}, \pmb{m}, \kappa$ introduced in section~\ref{sec:BGMM}, is based on \cite{RM11} for determination of $\pmb{\upalpha}$, and on \cite{RG97} for others.
\begin{figure}[h]
  \centering
  \includegraphics[width=\linewidth]{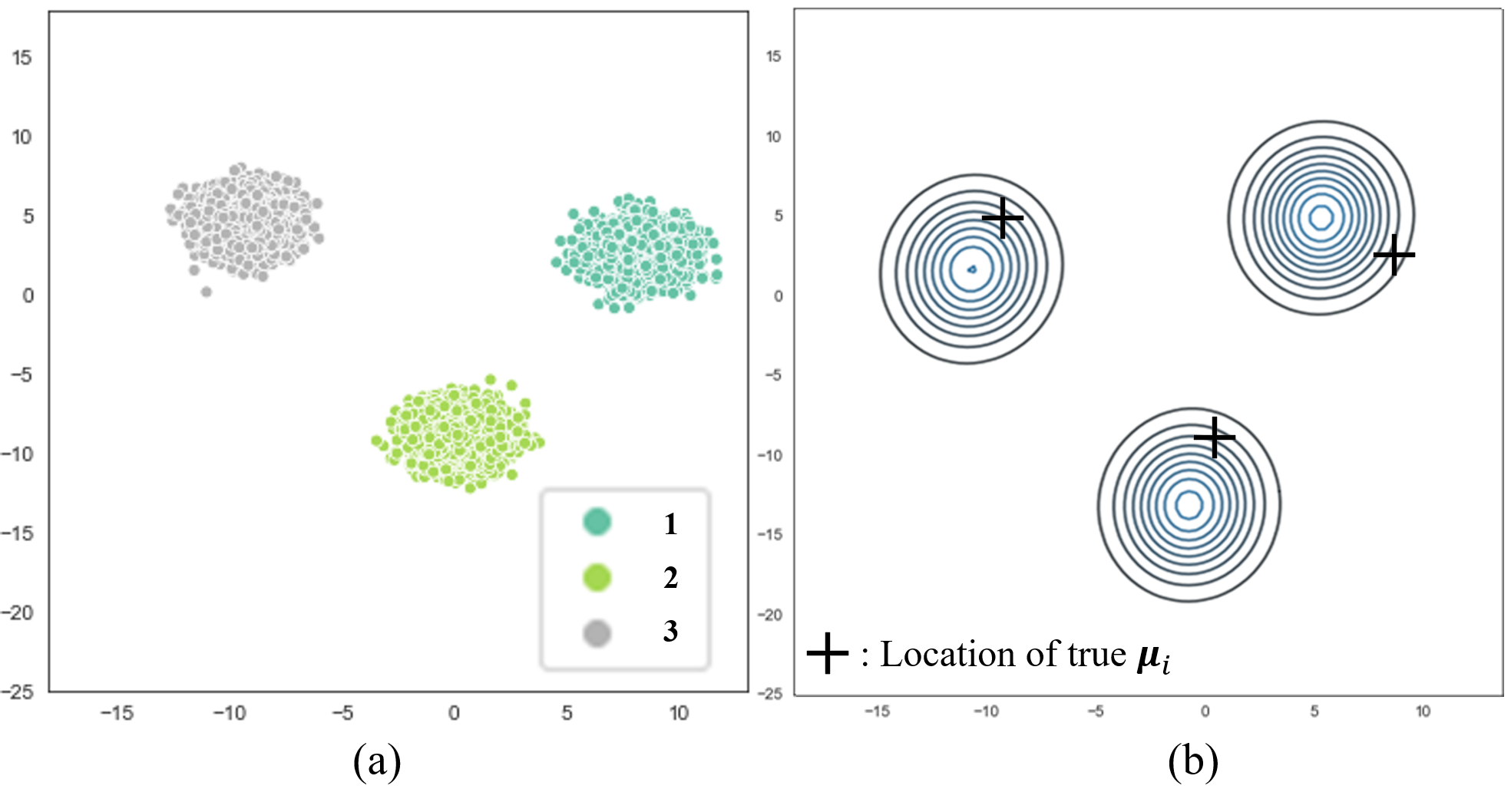}
  \caption{(a) Simulated data and (b) result of our algorithm}
  \label{fig:SD}
\end{figure}
 
\textbf{ii) Experimental result:} Total 9000 samples are simulated from three different Gaussian distributions, each of which has 3000 samples. (Figure~\ref{fig:SD}-(a)) For evaluating the accuracy of the result of parameter estimation, we first fit the GMM model on the simulated data with three latent clusters. 
After splitting data in three different subsets, let three local sites have each of them. 
Then, we used our GRAFFL framework to train the GMM model at the CS. 
As a result, the mode of the estimated posterior distribution of parameters $\pmb{\mathrm{\uptheta}}^{GRAFFL}=\{ \pmb{\mathrm{\uppi}}, \pmb{\mathrm{\upmu}}, \pmb{\mathrm{\Sigma}}\}$ is fairly accorded with the true parameter set $\pmb{\mathrm{\uptheta}}^{*}$ (Figure~\ref{fig:SD}-(b)). 

\subsection{Boosting analysis in a local site: imbalanced data}
\textbf{i) Experimental set-up:} We used PhysioNet2012 dataset \cite{Si+12} to simulate a horizontally distributed setting. We randomly split the dataset into $3$ sites (i.e. $S=3$) such that each site satisfies two situations: (1) binary classification of imbalanced data, (2) data which is classified well when all data is available at one site, but is not in a local site.
The partitioned dataset of each site was again divided into training and test sets each in a stratified manner. The sufficient summary statistics of each training data is obtained from the encoder network of SuffiAE trained at each site. The encoding dimension $d$ (i.e. dimensions of a latent vector) can be determined as an arbitrary number satisfying $d \leq D$.
In this experiment, $d$ was set to be $24$. 
We used logistic regression and AUC as a base classifier and a performance metric respectively. 

\textbf{ii) Experimental result:} 
\begin{figure}[h]
  \centering
  \includegraphics[width=\linewidth]{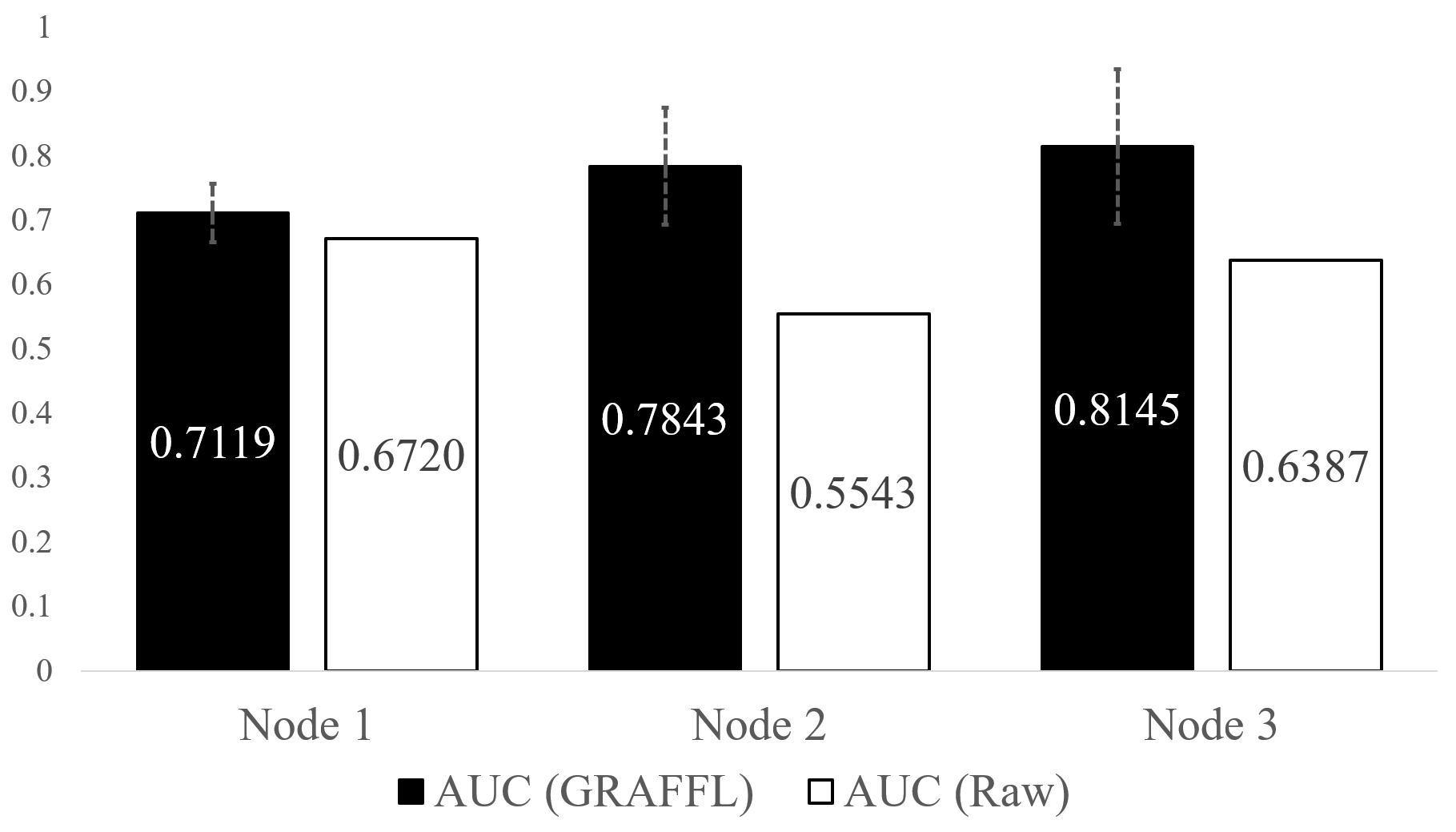}
  \caption{Summarized result on PhysioNet2012 dataset}
  \label{fig:SD}
\end{figure}
The detailed result is summarized in Table 1 of the Appendix.
Before splitting data into each site, we checked the performance of a classifier trained on all data ('All' in Table \ref{tbl:Physio}). This result is to be the upper bound of our proposed algorithm. In each site, we constructed a local classifier and tested the performance ('site $i$ Raw' in Table \ref{tbl:Physio}), which is to be the lower bound of our proposed algorithm.
After all, we over-sampled data of the minor class of each site using our algorithm ('site $i$ GRAFFL' in Table \ref{tbl:Physio}). We expected the over-sampling strategy to boost a classification performance in each site as the information of other site is reflected during the training of our algorithm.
We evaluated the classifier performance with different set of augmented samples more than 20 times, and averaged AUC score with standard deviation.
The result strongly implies that our algorithm can learn all the information with only exchanging minimal information (i.e. a discrepancy value) between the CS and each site, thereby can alleviate a problematic situation in analyzing a local data.

\subsection{Boosting analysis in a local site: scarce data}
\textbf{i) Experimental set-up: } Suppose the situation where extremely scarce data makes qualified data analyses be difficult. 
We used the Vehicle dataset \cite{Li+19}, which is already prepared for the horizontally distributed setting.
We set a situation where all six sites have a similar composition of data, while last three sites have lost many samples by accident. We denoted it as 'Vehicle $n\%$', which means that the last three sites have 
Therefore, a sample in a negative class (labeled as 1) cannot be classified well in the last three sites. (i.e. all the data is classified into true positive)
In that case, the three sites can derive benefit from GRAFFL framework as our algorithm can make an effect of collecting more samples instead for the last three sites.

\textbf{ii) Experimental result:} 

\begin{table}[]
\caption{Classification results of Vehicle dataset}
\label{tbl:Vehicle}
\begin{adjustbox}{width=0.48\textwidth}
\begin{tabular}{lllllll}
\hline
Vehicle(1\%)                       & \begin{tabular}[c]{@{}l@{}}Node 1\\ Raw\end{tabular} & \begin{tabular}[c]{@{}l@{}}Node 2\\ Raw\end{tabular} & \begin{tabular}[c]{@{}l@{}}Node 3\\ Raw\end{tabular}                      & \begin{tabular}[c]{@{}l@{}}Node 1\\ GRAFFL\end{tabular} & \begin{tabular}[c]{@{}l@{}}Node 2\\ GRAFFL\end{tabular} & \begin{tabular}[c]{@{}l@{}}Node 3\\ GRAFFL\end{tabular} \\ \hline
\multicolumn{1}{l|}{F1}            & \multicolumn{1}{l|}{0}                               & \multicolumn{1}{l|}{0}                               & \multicolumn{1}{l|}{0}                                                    & \multicolumn{1}{l|}{1}                                  & \multicolumn{1}{l|}{0.6667}                             & 1                                                       \\ \hline
\multicolumn{1}{l|}{Cut-off}       & \multicolumn{1}{l|}{0.8}                             & \multicolumn{1}{l|}{0.8}                             & \multicolumn{1}{l|}{0.8}                                                  & \multicolumn{1}{l|}{0.4168}                             & \multicolumn{1}{l|}{0.6053}                             & 0.503                                                   \\ \hline
\multicolumn{7}{l}{}                                                                                                                                                                                                                                                                                                                                                                                       \\
\hline \multicolumn{1}{l|}{Vehicle(5\%)}  & \begin{tabular}[c]{@{}l@{}}Node 1\\ Raw\end{tabular} & \begin{tabular}[c]{@{}l@{}}Node 2\\ Raw\end{tabular} & \multicolumn{1}{l|}{\begin{tabular}[c]{@{}l@{}}Node 3\\ Raw\end{tabular}} & \begin{tabular}[c]{@{}l@{}}Node 1\\ GRAFFL\end{tabular} & \begin{tabular}[c]{@{}l@{}}Node 2\\ GRAFFL\end{tabular} & \begin{tabular}[c]{@{}l@{}}Node 3\\ GRAFFL\end{tabular} \\ \hline
\multicolumn{1}{l|}{F1}            & \multicolumn{1}{l|}{0}                               & \multicolumn{1}{l|}{0}                               & \multicolumn{1}{l|}{0}                                                    & \multicolumn{1}{l|}{1}                                  & \multicolumn{1}{l|}{1}                                  & 0.9091                                                  \\ \hline
\multicolumn{1}{l|}{Cut-off}       & \multicolumn{1}{l|}{0.9524}                          & \multicolumn{1}{l|}{0.9524}                          & \multicolumn{1}{l|}{0.9375}                                               & \multicolumn{1}{l|}{0.5305}                             & \multicolumn{1}{l|}{0.3721}                             & 0.5077                                                  \\ \hline
                                   &                                                      &                                                      &                                                                           &                                                         &                                                         &                                                         \\
\hline\multicolumn{1}{l|}{Vehicle(10\%)} & \begin{tabular}[c]{@{}l@{}}Node 1\\ Raw\end{tabular} & \begin{tabular}[c]{@{}l@{}}Node 2\\ Raw\end{tabular} & \multicolumn{1}{l|}{\begin{tabular}[c]{@{}l@{}}Node 3\\ Raw\end{tabular}} & \begin{tabular}[c]{@{}l@{}}Node 1\\ GRAFFL\end{tabular} & \begin{tabular}[c]{@{}l@{}}Node 2\\ GRAFFL\end{tabular} & \begin{tabular}[c]{@{}l@{}}Node 3\\ GRAFFL\end{tabular} \\ \hline
\multicolumn{1}{l|}{F1}            & \multicolumn{1}{l|}{0}                               & \multicolumn{1}{l|}{0}                               & \multicolumn{1}{l|}{0}                                                    & \multicolumn{1}{l|}{1}                                  & \multicolumn{1}{l|}{0.9697}                             & 1                                                       \\ \hline
\multicolumn{1}{l|}{Cut-off}                            & \multicolumn{1}{l|}{0.9762}                                               & \multicolumn{1}{l|}{0.975}                                                & \multicolumn{1}{l|}{0.9667}                                                                    & \multicolumn{1}{l|}{0.9174}                                                  & \multicolumn{1}{l|}{0.9735}                                                  & 0.9619                                                  \\ \hline
\end{tabular}
\end{adjustbox}
\end{table}


As shown in the Table \ref{tbl:Vehicle}, it is impossible for the last three sites to classify the negative class ('site $i$ Raw' in Table \ref{tbl:Vehicle}), therefore the F1 score is zero. Nevertheless, after receiving boosting samples from the model trained by our algorithm made it possible to perfectly classify data of the negative label.
As a result, the cut-off value of each classifier is adjusted to other values, with increased F1 score close to 1.
It is therefore reasonable to conclude that our proposed algorithm can leverage the power of samples in other sites while no exchange of raw data.

\section{CONCLUSION}
This paper presents the first gradient-free FL framework for a Bayesian generative model and demonstrated its capability for practical applications by adopting the Bayesian GMM model. In the process of learning the federated Bayesian GMM via our framework, the only information to be transmitted between the local site and the CS is a discrepancy value between observed and generated data in the form of summary statistics. To guarantee our framework is feasible, we proposed SuffiAE which can provide sufficient summary statistics while preserving data privacy. Along with a theoretical analysis of the framework, experiments were conducted on synthetic and real datasets, which show that the proposed method is applicable in practices with the data-distributed environment. It can be further extended to other generative models such as GAN and will be served as a useful alternative to the gradient-based FL. 



\bibliographystyle{ACM-Reference-Format}
\bibliography{GRAFFL_ref}
\newpage
\appendix
\section{Appendix}

\subsection{Sufficiency proof of 3.3-(i) and Eq (1)}
At first, we need to find a lower bound of log-likelihood of joint density $\log{p(\pmb{\mathrm{x}}, \mathrm{y})}$.
\begin{multline}\notag
{\mathcal{L}(\pmb{\mathrm{x}}, \mathrm{y})} = \log{p(\pmb{\mathrm{x}}, \mathrm{y})} \\
=\log{\int_{\pmb{\mathrm{z}}}{p(\pmb{\mathrm{x}}|\pmb{\mathrm{z}})p({\mathrm{y}}|\pmb{\mathrm{z}})}p(\pmb{\mathrm{z}})} \mathrm{d}\pmb{\mathrm{z}}
=\log{\int_{\pmb{\mathrm{z}}}{p(\pmb{\mathrm{x}}|\pmb{\mathrm{z}})p({\mathrm{y}}|\pmb{\mathrm{z}})}
\frac{p(\pmb{\mathrm{z}})}{q_{\phi}(\pmb{\mathrm{z}}|\pmb{\mathrm{x}})}}q_{\phi}(\pmb{\mathrm{z}}|\pmb{\mathrm{x}}) \mathrm{d}\pmb{\mathrm{z}} \\
\geq 
\int_{\pmb{\mathrm{z}}}{\log\Bigg\{{p(\pmb{\mathrm{x}}|\pmb{\mathrm{z}})p({\mathrm{y}}|\pmb{\mathrm{z}})}
\frac{p(\pmb{\mathrm{z}})}{q_{\phi}(\pmb{\mathrm{z}}|\pmb{\mathrm{x}})}}\Bigg\}q_{\phi}(\pmb{\mathrm{z}}|\pmb{\mathrm{x}}) \mathrm{d}\pmb{\mathrm{z}} \text{ } (\because \text{Jensen's inequality}) \\
= \int_{\pmb{\mathrm{z}}}{\log{p(\pmb{\mathrm{x}}|\pmb{\mathrm{z}})}q_{\phi}(\pmb{\mathrm{z}}|\pmb{\mathrm{x}})} \mathrm{d}\pmb{\mathrm{z}}
+ \int_{\pmb{\mathrm{z}}}{\log{p({\mathrm{y}}|\pmb{\mathrm{z}})}q_{\phi}(\pmb{\mathrm{z}}|\pmb{\mathrm{x}})} \mathrm{d}\pmb{\mathrm{z}} \\
- \int_{\pmb{\mathrm{z}}}{\log{\frac{q_{\phi}(\pmb{\mathrm{z}}|\pmb{\mathrm{x}})}{p(\pmb{\mathrm{z}})}}q_{\phi}(\pmb{\mathrm{z}}|\pmb{\mathrm{x}})} \mathrm{d}\pmb{\mathrm{z}} \\
= E_{\pmb{\mathrm{z}} \sim q_{\phi}(\pmb{\mathrm{z}}|\pmb{\mathrm{x}})}[\log{p(\pmb{\mathrm{x}}|f_{\theta}(\pmb{\mathrm{z}}))}]+
E_{\pmb{\mathrm{z}} \sim q_{\phi}(\pmb{\mathrm{z}}|\pmb{\mathrm{x}})}[\log{p(\mathrm{y}|g_{\psi}(\pmb{\mathrm{z}}))}]\\ -\operatorname{KL}[q_{\phi}(\pmb{\mathrm{z}}|\pmb{\mathrm{x}}) || p(\pmb{\mathrm{z}})] \\
:= LB(\phi)
\end{multline}

Then, again for $\log{p(\pmb{\mathrm{x}}, \mathrm{y})}$,
\begin{multline}\notag
\log{p(\pmb{\mathrm{x}}, \mathrm{y})} = \log{p(\pmb{\mathrm{x}}, \mathrm{y})} \int_{\pmb{\mathrm{z}}}{q_{\phi}(\pmb{\mathrm{z}}|\pmb{\mathrm{x}}) \mathrm{d}\pmb{\mathrm{z}}}
=\int_{\pmb{\mathrm{z}}}{\log{p(\pmb{\mathrm{x}}, \mathrm{y})}q_{\phi}(\pmb{\mathrm{z}}|\pmb{\mathrm{x}}) \mathrm{d}\pmb{\mathrm{z}}} \\
=\int_{\pmb{\mathrm{z}}}{\log\Big\{{p(\pmb{\mathrm{x}}|\pmb{\mathrm{z}})p({\mathrm{y}}|\pmb{\mathrm{z}})}
p(\pmb{\mathrm{z}})}\Big\}q_{\phi}(\pmb{\mathrm{z}}|\pmb{\mathrm{x}}) \mathrm{d}\pmb{\mathrm{z}}\\
=\int_{\pmb{\mathrm{z}}}{\log\Big\{{p(\pmb{\mathrm{x}},\pmb{\mathrm{z}})p({\mathrm{y}}|\pmb{\mathrm{z}})}}\Big\}q_{\phi}(\pmb{\mathrm{z}}|\pmb{\mathrm{x}}) \mathrm{d}\pmb{\mathrm{z}}\\
=\int_{\pmb{\mathrm{z}}}{\log\Bigg\{{\frac{p(\pmb{\mathrm{x,z}})}{q_{\phi}(\pmb{\mathrm{z}}|\pmb{\mathrm{x}})}q_{\phi}(\pmb{\mathrm{z}}|\pmb{\mathrm{x}})p({\mathrm{y}}|\pmb{\mathrm{z}})}}\Bigg\}q_{\phi}(\pmb{\mathrm{z}}|\pmb{\mathrm{x}}) \mathrm{d}\pmb{\mathrm{z}} \\
=\int_{\pmb{\mathrm{z}}}{\log\Bigg\{{\frac{p(\pmb{\mathrm{x,z}})}{q_{\phi}(\pmb{\mathrm{z}}|\pmb{\mathrm{x}})}p({\mathrm{y}}|\pmb{\mathrm{z}})}}\Bigg\}q_{\phi}(\pmb{\mathrm{z}}|\pmb{\mathrm{x}}) \mathrm{d}\pmb{\mathrm{z}}
+ \int_{\pmb{\mathrm{z}}}{q_{\phi}(\pmb{\mathrm{z}}|\pmb{\mathrm{x}})\log{\{q_{\phi}(\pmb{\mathrm{z}}|\pmb{\mathrm{x}})\}}} \mathrm{d}\pmb{\mathrm{z}} \\
= LB(\phi) - H(q_{\phi}(\pmb{\mathrm{z}}|\pmb{\mathrm{x}}))
\end{multline}
where $H(\cdot)$ is Shannon's entropy function.

Let us define each term in $LB(\phi)$ as follows:
\begin{table}[hbt!]\notag
\begin{tabular}{|l|l|}
\hline
 $E_{\pmb{\mathrm{z}} \sim q_{\phi}(\pmb{\mathrm{z}}|\pmb{\mathrm{x}})}[\log{p(\pmb{\mathrm{x}}|f_{\theta}(\pmb{\mathrm{z}}))}]$ & Reconstruction error \\ \hline
 $E_{\pmb{\mathrm{z}} \sim q_{\phi}(\pmb{\mathrm{z}}|\pmb{\mathrm{x}})}[\log{p(\mathrm{y}|g_{\psi}(\pmb{\mathrm{z}}))}]$ & Classification error \\ \hline
 $\operatorname{KL}[q_{\phi}(\pmb{\mathrm{z}}|\pmb{\mathrm{x}}) || p(\pmb{\mathrm{z}})]$& Regularization       \\ \hline
\end{tabular}
\end{table}

Applying this to the setting of GRAFFL scheme, we at first have following objective function at each site $j$: \\
\begin{multline}\notag
\sum_{i=1}^{n_{j}} \log p(\pmb{\mathbf{x}}_{i}, y_{i}) \geq 
\sum_{i=1}^{n_{j}}\{E_{z\sim q_{\phi}(\pmb{\mathbf{z}} | \pmb{\mathbf{x}})}[\log p(\pmb{\mathbf{x}}_{i} | f_{\theta}(\pmb{\mathbf{z}}))] \\
+ E_{\pmb{\mathbf{z}} \sim q_{\phi}(\pmb{\mathbf{z}} | \pmb{\mathbf{x}})}[\log p(y_{i} | g_{\psi}(\pmb{\mathbf{z}}))] -\operatorname{KL}[q_{\phi}(\pmb{\mathbf{z}} | \pmb{\mathbf{x}}_{i})|| p(\pmb{\mathbf{z}}_{i})]\}
\end{multline}

Then, for reconstruction error term,
\begin{multline}\notag
E_{z\sim q_{\phi}(\pmb{\mathbf{z}} | \pmb{\mathbf{x}})}[\log p(\pmb{\mathbf{x}}_{i} | f_{\theta}(\pmb{\mathbf{z}}))]\\
=\int_{\pmb{\mathrm{z}}}{\log{p(\pmb{\mathbf{x}}_{i}|f_{\theta}(\pmb{\mathbf{z}}))q_{\phi}(\pmb{\mathrm{z}}|\pmb{\mathrm{x}}_{i})}} \mathrm{d}\pmb{\mathrm{z}}
\cong 
\log{p(\pmb{\mathbf{x}}_{i}|f_{\theta}(\pmb{\mathbf{z}}_{i})})\\
=\log \prod_{k=1}^{D} p(\pmb{\mathbf{x}}_{i, k} | f_{\theta}(\pmb{\mathbf{z}}_{i})) 
=\sum_{k=1}^{D}{\log{p(\pmb{\mathbf{x}}_{i, k} | f_{\theta}(\pmb{\mathbf{z}}_{i}))}} \\
\propto -\frac{1}{2}\sum_{k=1}^{D}{(\pmb{\mathbf{x}}_{i,k}-\pmb{\mathbf{x}}_{i,k}^{'})^2} \\
\end{multline}
, where  
\begin{multline}\notag
\pmb{\mathbf{x}}_{i}^{'}=f_{\theta}(\pmb{\mathbf{z}}_{i})=\text{Decoder}_{\theta}({\pmb{\mathbf{x}}^{\text{enc}}_{i}+\pmb{\mathbf{x}}^{\text{noise}}_{i}}), \pmb{\mathbf{x}}^{\text{noise}}_{i} \sim \mathcal{N}(\pmb{0}, \upalpha \pmb{\mathbf{I}})
\end{multline}

Next, for classification error term,
\begin{multline}\notag
E_{\pmb{\mathbf{z}} \sim q_{\phi}(\pmb{\mathbf{z}} | \pmb{\mathbf{x}})}[\log p(y_{i} | g_{\psi}(\pmb{\mathbf{z}}))]
=\int_{\pmb{\mathrm{z}}}{\log{p({\mathbf{y}}_{i}|g_{\psi}(\pmb{\mathbf{z}}))q_{\phi}(\pmb{\mathrm{z}}|\pmb{\mathrm{x}}_{i})}} \mathrm{d}\pmb{\mathrm{z}}\\
\propto \mathbf{y}_{i}\log{g_{\psi}(\pmb{\mathbf{z}}_{i}))}+(1-\mathbf{y}_{i})\log(1-{g_{\psi}(\pmb{\mathbf{z}}_{i}))})
\end{multline}
, where $g_{\psi}(\cdot)$ is a logistic sigmoid function.

Last, for regularization term $\operatorname{KL}[q_{\phi}(\pmb{\mathbf{z}} | \pmb{\mathbf{x}}_{i})|| 
p(\pmb{\mathbf{z}}_{i})]$,
\begin{multline}\notag
\operatorname{KL}[q_{\phi}(\pmb{\mathbf{z}} | \pmb{\mathbf{x}}_{i})|| p(\pmb{\mathbf{z}}_{i})]\\
=\frac{1}{2}\{\Tr(\upalpha \pmb{\mathbf{I}})+q_{\phi}(\pmb{\mathbf{x}}_{i})^\intercal q_{\phi}(\pmb{\mathbf{x}}_{i})-\operatorname{dim}{(q_{\phi}(\pmb{\mathbf{x}}_{i}))}-\operatorname{dim}{(q_{\phi}(\pmb{\mathbf{x}}_{i}))}\log{\upalpha}\} \\
=\Bigg(\frac{1}{2}\sum_{m=1}^{d}(\upalpha -1-\log{\upalpha})+\|q_{\phi}(\pmb{\mathbf{x}}_{i})\|_{2}^{2}\Bigg)
\end{multline}
By summing up all three terms together, we can get Eq (1).

\subsection{Detailed results on PhysioNet2012 dataset}
\begin{table}[!h]
\caption{Classification results of PhysioNet2012 dataset}
\label{tbl:Physio}
\begin{adjustbox}{width=0.50\textwidth}
\begin{tabular}{c|c|ccc|ccc}
\hline PhysioNet2012                                                           & All    & \begin{tabular}[c]{@{}c@{}}site 1\\ Raw\end{tabular} & \begin{tabular}[c]{@{}c@{}}site 2\\ Raw\end{tabular} & \begin{tabular}[c]{@{}c@{}}site 3\\ Raw\end{tabular} & \begin{tabular}[c]{@{}c@{}}site 1\\ GRAFFL\end{tabular} & \begin{tabular}[c]{@{}c@{}}site 2\\ GRAFFL\end{tabular} & \begin{tabular}[c]{@{}c@{}}site 3\\ GRAFFL\end{tabular} \\ \hline
Class Ratio                                                             & 6:1    & \multicolumn{1}{c|}{5:1}                             & \multicolumn{1}{c|}{5:1}                             & 6:1                                                  & \multicolumn{1}{c|}{1:1}                                & \multicolumn{1}{c|}{1:1}                                & 1:1                                                     \\ \hline
\begin{tabular}[c]{@{}c@{}}AUC\\ (Average)\end{tabular}                 & 0.8158 & \multicolumn{1}{c|}{0.6720}                          & \multicolumn{1}{c|}{0.5543}                          & 0.6387                                               & \multicolumn{1}{c|}{0.7119}                             & \multicolumn{1}{c|}{0.7843}                             & 0.8145                                                  \\ \hline
Cut-off                                                                 & 0.2313 & \multicolumn{1}{c|}{0.7169}                          & \multicolumn{1}{c|}{0.7874}                          & 0.7111                                               & \multicolumn{1}{c|}{0.2594}                             & \multicolumn{1}{c|}{0.6056}                             & 0.4072                                                  \\ \hline
\begin{tabular}[c]{@{}c@{}}Standard \\ Deviation \\ of AUC\end{tabular} & -      & \multicolumn{1}{c|}{-}                               & \multicolumn{1}{c|}{-}                               & -                                                    & \multicolumn{1}{c|}{0.0772}                             & \multicolumn{1}{c|}{0.0909}                             & 0.1198                                                  \\ \hline
\end{tabular}
\end{adjustbox}
\end{table}

\end{document}